\documentclass[10pt,twocolumn,letterpaper]{article}

\usepackage{iccv}
\usepackage{times}
\usepackage{epsfig}
\usepackage{graphicx}
\usepackage{amsmath}
\usepackage{amssymb}
\usepackage{multirow}
\usepackage{cite}
\usepackage{color}
\usepackage{float}
\usepackage[table,dvipsnames]{xcolor}
\usepackage[pagebackref=true,breaklinks=true,letterpaper=true,colorlinks,bookmarks=false]{hyperref}
\iccvfinalcopy 
\def\iccvPaperID{***} 

\ificcvfinal\fi

\title{Lite-FPN for Keypoint-based Monocular 3D Object Detection}
\author{%
	Lei Yang\textsuperscript{1},  Xinyu Zhang\textsuperscript{1}\thanks{Corresponding author: xyzhang@tsinghua.edu.cn},  Li Wang\textsuperscript{1},  Minghan Zhu\textsuperscript{2},  Jun Li\textsuperscript{1} \\
	\textsuperscript{1}State Key Laboratory of Automotive Safety and Energy,  Tsinghua University\\
	\textsuperscript{2}University of Michigan\\
	\texttt{\{yanglei20@mails., xyzhang@, wangli\_thu@mail., lijun1958@\}tsinghua.edu.cn} \\
	\texttt{minghanz@umich.edu}
}
\ificcvfinal\fi

\begin{document}
    \maketitle
    \begin{abstract}
    3D object detection with a single image is an essential and challenging task for autonomous driving. Recently, keypoint-based monocular 3D object detection has made tremendous progress and achieved great speed-accuracy trade-off. However, there still exists a huge gap with LIDAR-based methods in terms of accuracy. To improve their performance without sacrificing efficiency, we propose a sort of lightweight feature pyramid network called Lite-FPN to achieve multi-scale feature fusion in an effective and efficient way, which can boost the multi-scale detection capability of keypoint-based detectors. Besides, the misalignment between classification score and localization precision is further relieved by introducing a novel regression loss named attention loss. With the proposed loss, predictions with high confidence but poor localization are treated with more attention during the training phase. Comparative experiments based on several state-of-the-art keypoint-based detectors on the KITTI dataset show that our proposed methods manage to achieve significant improvements in both accuracy and frame rate. The code and pretrained models will be released at \url{https://github.com/yanglei18/Lite-FPN}.
    \end{abstract}

    \section{Introduction}
    \label{sec:introduction}
	3D object detection responsible for providing precise 3D bounding boxes of surrounding objects is an essential environmental perception task in autonomous driving. Recently, relying on the accurate depth measurements of LIDAR, LIDAR-based detectors have achieved superior performance. However, several intrinsic deficiencies of LIDAR system such as high cost and sensitivity to adverse weather conditions inevitably impose limitations on the applications of these methods. In contrast, camera sensors are more economical, robust in rain and snow, and feasible to satisfy the strict vehicle regulations.

    \begin{figure}[t]
    	\centering
    	\includegraphics[width=4.1cm]{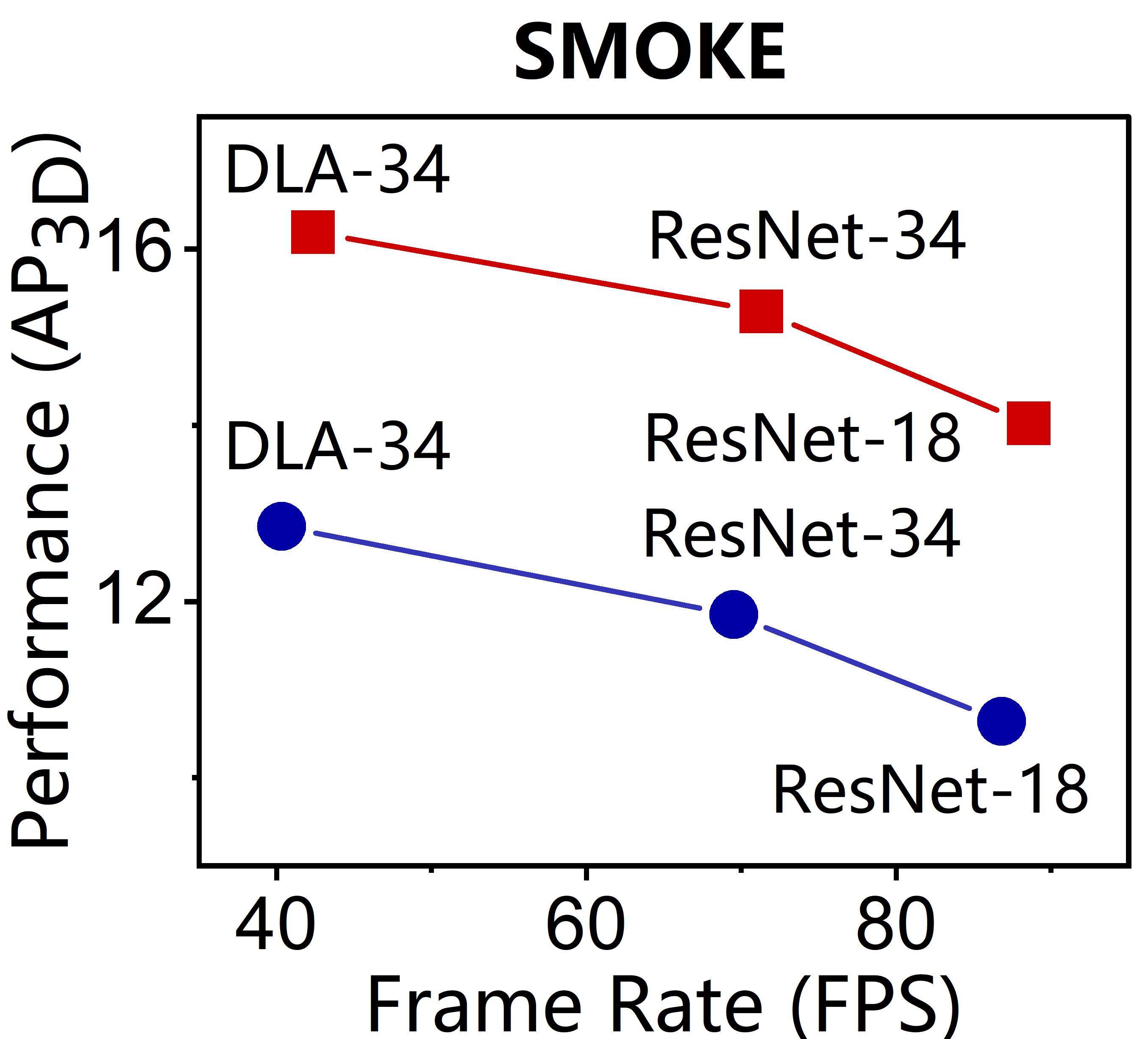}
    	\includegraphics[width=4.1cm]{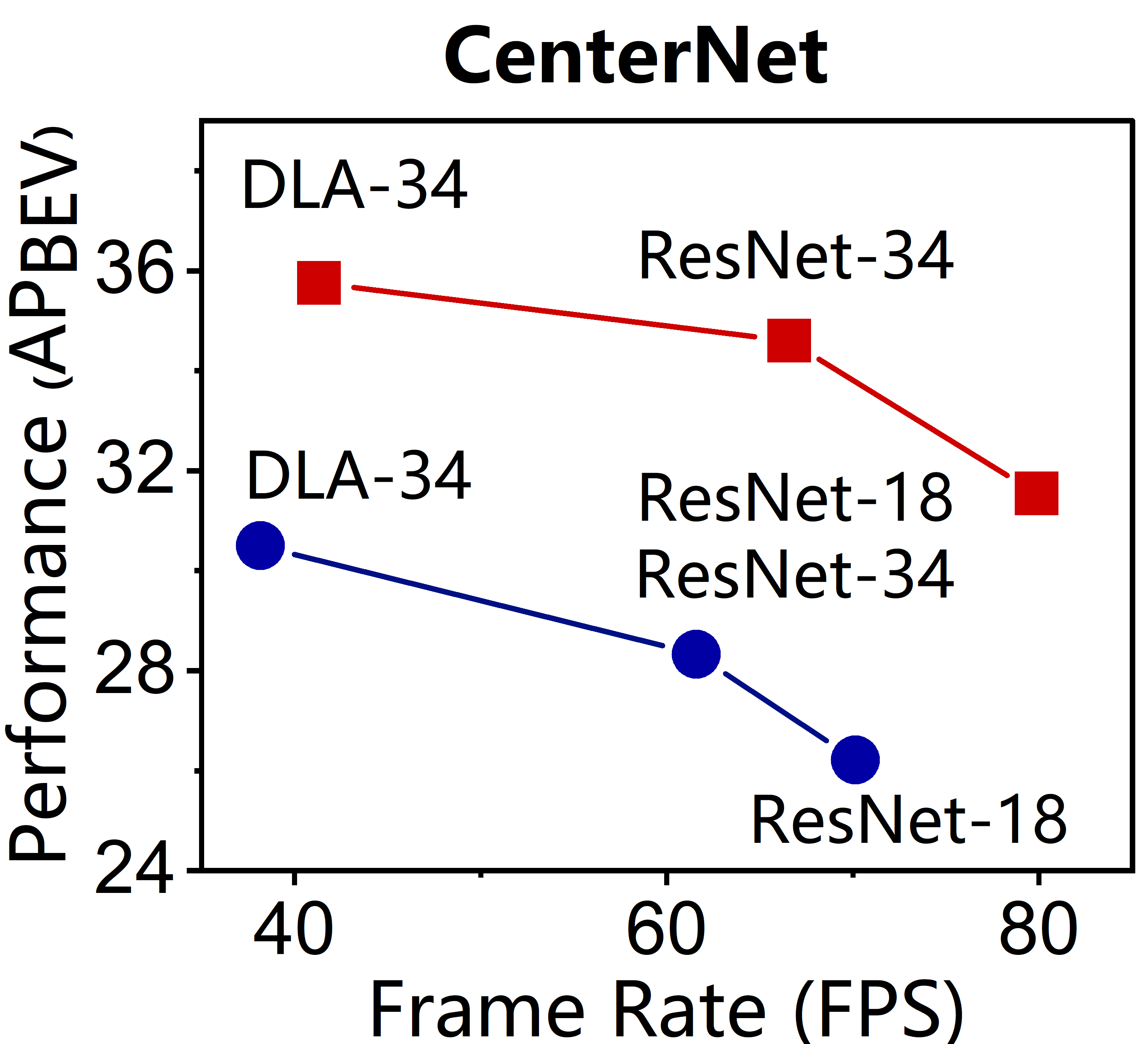}
    	\setlength{\abovecaptionskip}{5pt}
		\setlength{\belowcaptionskip}{-10pt}
    	\caption{Detection performance vs speed for two baselines and their  improved models with our proposed Lite-FPN module and attention loss. The results of baselines are drawn as blue dots, and red squares represent the results of improved models. The average precision of 3D IoU with an IoU threshold of 0.7 is selected to evaluate SMOKE\cite{liu2020smoke} related models. The evaluation metric for CenterNet\cite{zhou2019objects} related methods is the average precision for bird's-eye view with an IoU threshold of 0.5. All results are evaluated on the KITTI validation set. The improved models with our proposed methods outperform their baselines in terms of both speed and accuracy by a massive margin.}
    \label{fig:one}
    \end{figure}
    
    According to ways of obtaining 3D information, monocular 3D object detectors can be roughly divided into four groups: pseudo-LiDAR, 2D/3D geometry constraint, anchor-based and keypoint-based methods. Pseudo-LiDAR related methods \cite{wang2019pseudo,you2019pseudo,Ku2019Monocular3O,vianney2019refinedmpl,weng2019monocular,ma2019accurate,manhardt2019roi,wang2021depth,reading2021categorical}acquire depth image using the existing monocular depth estimation algorithm\cite{klingner2020self,qiao2020vip,aich2020bidirectional} and transform the acquired depth image to pseudo-LiDAR point cloud which can be fed to LIDAR-based 3D object detectors\cite{lang2019pointpillars,yang20203dssd,he2020structure,shi2020pv,zheng2021se}. However, these methods usually suffer from poor efficiency due to the additional depth estimation module. Other methods based on 2D/3D geometry constraint \cite{mousavian20173d,liu2019deep,choi2019multi,li2019gs3d,chabot2017deep} infer translation information on the basis of the geometric relationship between 2D and 3D properties on the image plane. Owing to the sensitiveness to inaccurate prior properties such as 2D boxes, it is difficult for these methods to achieve high accuracy and low latency simultaneously.Anchor-based frameworks \cite{brazil2019m3d,qin2019triangulation,chen2016monocular,luo2021m3dssd}obtain 3D bounding boxes by scoring and refining the predefined dense 3D anchor boxes. The speed of these methods is heavily restricted by the densely sampled anchors. Keypoint-based detectors \cite{zhou2019objects,tang2020center3d,li2020rtm3d,chen2020monopair,shi2020distance,liu2020smoke,zhang2021objects,ma2021delving} directly estimate all the 3D properties of instance at keypoint location. Compared with methods in other groups, detectors in keypoint-based group have achieved great speed-accuracy balance and thus are more appropriate for the autonomous driving applications. However, there still exists a large gap in detection precision between monocular-based detectors and LIDAR-based methods. In this paper, we aim to improve the accuracy of keypoint-based methods and meanwhile strengthen their speed advantage.
    	
    One key challenge for keypoint-based monocular 3D object detectors is to handle objects within a large range of scales and distances. Multi-scale detection requires sufficient multi-scale information, yet most keypoint-based detectors merely rely on a high-resolution feature map. In this paper, we achieve multi-scale feature fusion by introducing a kind of lightweight feature pyramid network called Lite-FPN which is a generic module and can be integrated into most keypoint-based methods. In the proposed module, we firstly sample features from the feature maps in different resolutions via the pixel indices of candidate keypoints. Then, the sampled features are concatenated together for the following regression task. Compared with the origin regression layer applying on the whole high-resolution feature map, taking regression merely at meaningful keypoints is much more efficient. The time consumption introduced by the additional sampling and concatenation operations in Lite-FPN is negligible. Consequently, with the proposed Lite-FPN module, we achieve a more accurate and efficient monocular 3D object detection.
    	
    Another key challenge for keypoint-based monocular 3D object detectors is to alleviate the misalignment between classification scores and localization precision. Detections with high classification scores but low IoU with ground truth are prioritized over those with low scores but high IoU during evaluation, which leads to a lower AP at high IoU threshold. To tackle this problem, we propose a simple but effective regression loss named attention loss, in which predictions with high scores but poor localization are treated with more attention. Accordingly, boxes with high confidence can be better localized under this new training strategy.

    Our contributions are summarized as follows:
    \begin{itemize}
    	\item We propose a generic Lite-FPN module that conducts multi-scale feature fusion for keypoint-based monocular 3D object detectors in an efficient way, which is essential to handle objects within a large range of scales and distances.
    	\item A novel attention loss is proposed to solve the misalignment between classification score and localization precision without sacrificing detectors' efficiency.
    	\item By integrating the Lite-FPN module and the attention loss into several state-of-the-art keypoint-based detectors, their effectiveness has been verified on the public KITTI object 3D and BEV detection benchmarks.
    \end{itemize}
    
    \begin{figure*}
    \begin{center}
    \includegraphics[width=0.95\textwidth]{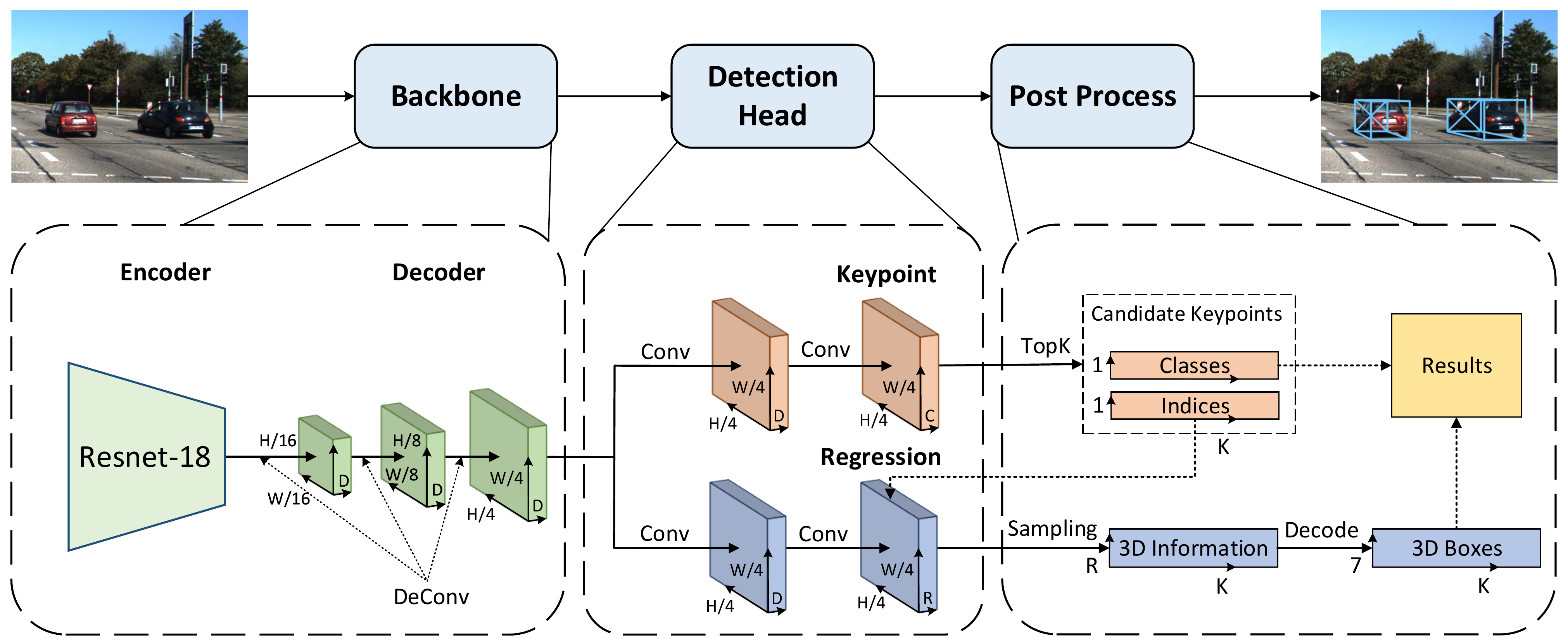}
    \end{center}
       \caption{The base architecture of a keypoint-based monocular 3D object detector.}
    \label{fig:two}
    \end{figure*}
    
    \section{Related work}
    \label{sec:related_work}
    {\bf Monocular 3D Object Detectors.} 
    Reconstructing spatial information is the core issue for monocular 3D object detection. Some methods \cite{wang2019pseudo,you2019pseudo,Ku2019Monocular3O,vianney2019refinedmpl,weng2019monocular,ma2019accurate,manhardt2019roi,wang2021depth,reading2021categorical} rely on the existing monocular depth estimation algorithms\cite{klingner2020self,qiao2020vip,aich2020bidirectional}. Pseudo-LIDAR \cite{wang2019pseudo} transforms the estimated depth image to artificial dense point clouds so as to employ LIDAR-based 3D object detectors \cite{ku2018joint,qi2018frustum}. Pseudo-LIDAR++\cite{you2019pseudo} extends the idea of Pseudo-LIDAR by refining the depth image using sparse point clouds from low-cost lidar. Another methods \cite{mousavian20173d,liu2019deep,choi2019multi,li2019gs3d,chabot2017deep} infer 3D information using 2D/3D geometry constraint. Deep3DBox \cite{mousavian20173d} obtains 3D translation information by solving the linear equations based on the geometry constraint between 2D box and projected 3D box on the image plane. Deep MANTA\cite{chabot2017deep} predicts some properties like 2D bounding box, category, 2D keypoints and the template similarity which is used to select the best matching 3D CAD model. Then, the orientation and 3D location are recovered through 2D/3D matching process. Anchor-based methods \cite{brazil2019m3d,qin2019triangulation,chen2016monocular,luo2021m3dssd} get location information using the predefined anchor boxes in 3D space. Mono3D \cite{chen2016monocular} generates 3D anchor boxes in dense distribution on the basis of ground plane assumption. Then it refines and scores each candidate anchor using handcraft features and performs NMS operation to obtain the final results. The remaining keypoint-based detectors\cite{zhou2019objects,tang2020center3d,li2020rtm3d,chen2020monopair,shi2020distance,liu2020smoke,zhang2021objects,ma2021delving} directly estimate the 3D properties of instance based on the high-dimensional features at keypoint position. CenterNet\cite{zhou2019objects} is the pioneering work of this group which takes the center of 2D box as keypoint and regresses all other 3D object properties such as depth, dimension and orientation directly at keypoint location.
    
    {\bf Multi-scale Detection.} 
    For keypoint-based 3D object detectors, adopting multi-scale features in detection head is effective to detect objects within a large range of scales and distances. 
    
    Referring to the FPN architecture in 2D object detection, UR3D \cite{shi2020distance} proposes a multi-scale framework to learn an unified representation for objects with different scale and distance properties. In this framework, five different detection heads sharing learnable weights are applied on five feature maps in different resolutions, objects are assigned onto different heads according to their scales and distances. Intuitively, this proposed multi-scale framework in UR3D\cite{shi2020distance} would be helpful to improve detection precision, but the additional detection heads and unnecessary NMS operation also lead to increased latency. RTM3D\cite{li2020rtm3d} presents a keypoint feature pyramid network (KFPN) structure to generate the feature maps containing multi-scale information. However, the additional upsampling and weighted sum operations inevitably increase the inference time of detector.
    	
    In comparison to the above works, our proposed Lite-FPN module achieves multi-scale feature fusion in a more efficient and effective way. All detections are predicted by a single detection head, which avoids the additional non-maximum suppression (NMS) process.
    
    {\bf Misalignment between Classification Score and Localization Precision.} 
    For detection results, alleviating the gap between their classification scores and localization precision is an effective way to improve the average precision of detectors.
    
    FCOS \cite{tian2019fcos} adds an additional branch to predict the center-ness which is correlated with localization precision. The final score of a bounding box is the product of classification score and the predicted center-ness. KM3D \cite{li2020monocular} introduces a 3D confidence branch to estimate the 3D IoU between the regressed box and ground truth. The final confidence of a bounding box is the down-weighted classification score by the estimated 3D IoU. Although these innovations can improve the consistency between the final confidence score and localization precision in some degree, the auxiliary branch will lead to poor real-time efficiency.
    	
    Compared with the previous works, our proposed attention loss focuses on optimizing the bounding boxes with high confidence but poor localization by treating these boxes with more attention in regression loss, which in return alleviates the misalignment between classification score and localization precision without hurting efficiency.
    
    \section{Method}
    \label{sec:method}
    \subsection{Keypoint-based Monocular 3D Object Detector}
    \label{sec:keypoint-based_monocular_3d_object_detector}
    The seminal work of keypoint-based monocular 3D object detectors is CenterNet \cite{zhou2019objects}. Recent works like SMOKE \cite{liu2020smoke} and Center3D\cite{tang2020center3d} retain the main architecture of CenterNet\cite{zhou2019objects} and achieve better performance by applying further optimization.
	
	Here we introduce the major structure shared by the detectors mentioned above. As shown in Figure.~\ref{fig:two}, the whole framework can be divided into four parts: (1) backbone, (2) detection head, (3) post process, (4) loss function.\\
	{\bf Backbone.} The backbone network is composed of an encoder and a decoder. The encoder extracts high-dimensional features from a RGB image with residual networks (ResNet)\cite{he2016deep} or deep layer aggregation (DLA-34) \cite{yu2018deep}. The decoder upsamples the bottleneck features to 1/4 times with respect to the input image by three deconvolutional layers.\\
	{\bf Detection Head.} The detection head consists of two task-specific branches: a keypoint branch and a regression branch. They are both formed by fully convolutional layers. The keypoint branch predicts the existence probability and class of object's keypoint at each pixel. The export heatmap from keypoint branch can be defined as $Y^{'}\in[0,1]^{H/4\times W/4\times C}$, where C is the number of object classes. The regression branch predicts the geometric parameters of 3D bounding box. The output regression map from the second branch can be expressed as $S^{r}\in\mathbb{R}^{H/4\times W/4\times R}$, where R denotes the number of encoded parameters indicating 3D information.\\
	{\bf Post Process.} The post process contains two sequence processes: a top-K process and a decode process. The top-K operation is applied on the heatmap to get a certain amount of candidate keypoints according to the descending order of keypoint existence probability. The acquired candidate keypoints can be represented as pixel indices and classes. The decode process is applied to the 3D information sampled from regression map via the above pixel indices to obtain the final 3D bounding boxes.

    \begin{figure*}
    \begin{center}
    \includegraphics[width=0.95\textwidth]{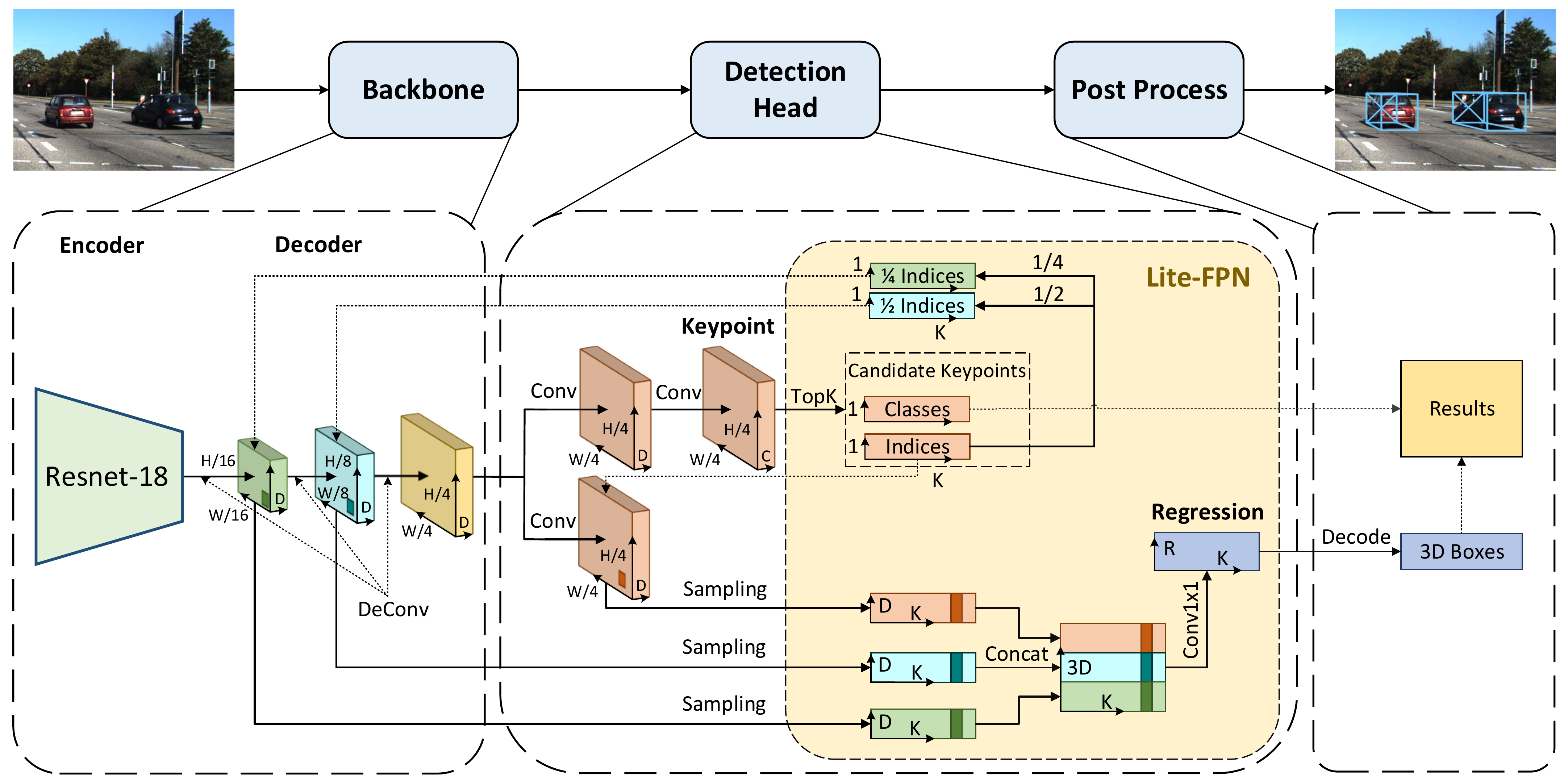}
    \end{center}
       \caption{Detector Overview. The main components of the improved detector are backbone, detection head and post process. See Section~\ref{sec:lite-fpn} for more details. We leverage the backbone network to extract multi-scale feature maps from an input image. In detection head, the keypoint branch is attached to the largest scale feature map to perform keypoint estimation. The regression layer is attached to the embedding features in Lite-FPN module to perform 3D box regression. The detection results are obtained via post process using the information from detection head.}
    \label{fig:three}
    \end{figure*}
    
    {\bf Loss Function.} The total loss function is made up of two parts: a keypoint classification loss and a regression loss.
	
	The keypoint classification loss is applied on the whole heatmap $Y^{'}\in[0,1]^{H/4\times W/4\times C}$. The ground truth heatmap $Y\in[0,1]^{H/4\times W/4\times C}$ can be produced by applying Gaussian kernel $Y_{cxyp}=e^{-\frac{(x-\tilde{p}_{x})^2 + (y-\tilde{p}_{y})^2}{2\sigma_p^2}}$ on each ground truth keypoint $p\in\mathbb{R}^2$ of class $c$, where $\sigma_p$ is the standard deviation related to the object size on image plane \cite{zhou2019objects,law2018cornernet}, $\tilde{p}=\left\lfloor\frac{p}{4}\right\rfloor$ is the counterpart of ground truth keypoint $p$ on the down-sampled heatmap. The final values on the ground truth heatmap $Y\in[0,1]^{H/4\times W/4\times C}$ are computed by an element-wise maximum operation $Y_{cxy} = \max \limits_{p}Y_{cxyp}$. The classification loss is defined as Eq.~\ref{con:eq1} based on a revised focal loss \cite{zhou2019objects,lin2017focal}.
    $$
	L_k=-\frac{1}{N}\sum_{c}^{C}\sum_{x}^{\frac{W}{4}}\sum_{y}^{\frac{H}{4}}
	$$
	\begin{equation}
		\begin{cases} 
			(1-Y_{cxy}^{'})^{\alpha}log(Y_{cxy}^{'})  & if \ Y_{cxy}=1\\
			(1-Y_{cxy})^\beta(Y_{cxy}^{'})^{\alpha}log(1-Y_{cxy}^{'}) & otherwise
		\end{cases},
		\label{con:eq1}
	\end{equation}
	
	where $\alpha$ and $\beta$ are two hyper-parameters, $N$ represents the number of keypoints per batch images. The term $(1-Y_{cxy})^{\beta}$ is responsible for alleviating the imbalance between positive and negative samples by downweighting the negative ones closed to ground truth keypoints.
	
	The regression loss is only applied on part of the regression map $S_{r}$. We define the objective function as Eq.~\ref{con:eq2} based on $l_1$ norm.
	\begin{equation}
		L_{reg}=\frac{1}{N}\sum_{i=0}^{N}l_{reg_{i}}=\frac{1}{N}\sum_{i=0}^{N}
		\begin{Vmatrix}
			\tau_{i}^{'}-\tau_{i}
		\end{Vmatrix}_{1},
		\label{con:eq2}
	\end{equation}
	where $l_{reg_{i}}$ is the sub-regression loss of $i-th$ keypoint $p_i$, and $\tau_i^{'}$ is a R-tuple encoded 3D information sampled from the regression map $S^{r}$, and $\tau_{i}$ denotes the ground truth of $\tau_{i}^{'}$.
	
	The total objective cost function is the summation of keypoint classification loss and regression loss:
	\begin{equation}
		L=L_{k}+\lambda L_{reg} ,
		\label{con:eq3}
	\end{equation}
	where $\lambda$ is the hyper-parameter that controls the proportion of regression loss in the total loss.
    
    \subsection{Lite-FPN}
    \label{sec:lite-fpn}
	{\bf Multi-Scale Detection.} In the feature hierarchical layers of backbone network, the high-resolution feature maps possessing more accurately localized features and smaller receptive fields are adept at detecting small-scale objects. Conversely, the low-resolution maps providing stronger high-level semantics and larger receptive fields are appropriate for the objects in large scale \cite{lin2017feature}. For most keypoint-based detectors, the detection head composed of keypoint and regression subnets is applied on the largest scale feature maps in backbone as shown in Figure.~\ref{fig:two}. On account of the deficiency of multi-scale information on a single feature map, it is challenging for these detectors to cover all objects within a large range of scales and distances. Therefore, taking full advantage of multi-scale features in detection head is an effective way to improve the performance of keypoint-based detectors.
	
	On the basis of the experimental results in CenterNet \cite{zhou2019objects} and SMOKE \cite{liu2020smoke}, the performance of 2D detection has reached state of the art while the average precision of 3D detection is still far away from that of LIDAR-based detectors. In the standard architecture of keypoint-based detector, the keypoint branch is shared by both 2D and 3D detection tasks, which explains that the keypoint branch is not the bottleneck that hinders the performance of 3D object detection. For this reason, we should emphatically make the utmost use of multi-scale information on the regression branch.
	
	There exists plenty of redundant data on the regression map. In the training phase, only the 3D information sampled by the pixel indices of ground truth keypoints is selected to compute the regression loss. During the inference phase, only the predictions sampled by the pixel indices of candidate keypoints are applied to generate results. Therefore, applying multi-scale feature fusion on some essential pixels of regression map can be an efficient and effective optimization scheme.
	
	{\bf Our Solution.} Based on the above analysis, we present a class of lightweight feature pyramid network called Lite-FPN for keypoint-based monocular 3D object detectors. The improved framework integrating our proposed Lite-FPN module is shown in Figure.~\ref{fig:three}. This detector consists of a backbone network, a detection head, and a post process unit. Compared with existing keypoint-based detectors, the backbone network remains unchanged. The top-K operation is transferred from post process to the Lite-FPN module in detection head. The post process simply decodes the regression parameters to 3D bounding boxes. 
	
	In terms of the data flow, the newly designed regression branch with Lite-FPN module can be divided into three phases: (1) keypoint proposal phase, (2) feature fusion phase, (3) regression phase. In the first phase, candidate keypoints expressed as classes and pixel indices are obtained via a top-K operation applied on the heatmap. In the feature fusion phase, based on the proportionate relationship of the pixel indices on different feature maps, we can get the pixel indices of candidate keypoints on the two feature maps with 1/8 and 1/16 resolution of the input image by multiplying the previous pixel indices with 1/2 and 1/4 respectively. Then, three groups of features with the same shape $K \times D$ are sampled from the feature maps with 1/4, 1/8 and 1/16 resolution of the input image via the above three sets of pixel indices. Finally, the sampled features are concatenated together to create the embedding features with the shape of $K \times 3D$. In regression phase, the embedding features are fed to a $1 \times 1$ convolutional layer to predict the geometric parameters related to 3D bounding box.

    \subsection{Attention Loss}
    \label{sec:attention_loss}
	{\bf False Attention.} Owing to the independence between keypoint branch and regression branch in the original training phase, there is a misalignment between classification score and localization precision. It is common for keypoints to generate detection boxes with high confidence scores but poor localization, which tends to be false positives during evaluation, leading to a lower AP at a strict 3D IoU threshold.
	
	Under the traditional training strategy, all keypoints are treated with equal attention in Eq.~\ref{con:eq2}, which is the point of focus we can make further improvement. Since boxes are produced on keypoints locations, they share the same confidence score and localization precision. We can divide keypoints into four groups: high confidence but poor localization, high confidence and accurate localization, low confidence and poor localization, low confidence but accurate localization. Keypoints with high confidence have sufficient feature representation well aligned with instance to make great localization prediction. Hence, these keypoints should receive more attention than the ones with lower confidence. Among these high-confidence keypoints, some inaccurate ones are inclined to generate false positives and should be given more attention to get further improvements. The rest with accurate localization are more likely to produce true positives and are supposed to be given relatively less attention to maintain the current situation. On the other hand, keypoints with low confidence are considered as hard examples and easy to be filtered out during the post process. Some of these keypoints with poor localization can be focused in some degree to achieve some improvements. The others with accurate localization have reached the bottleneck and simply need to be given the least attention.
	
	In other words, different attention should be paid to different keypoints in regression function. It is indispensable to focus on optimizing the noteworthy keypoints.
	
	{\bf Our Solution.} To alleviate the above false attention issue, we propose a simple but effective attention loss as Eq.~\ref{con:eq4}. The basic idea of the proposed loss is to assign an attention weight $w_i$ to the sub-item $l_{reg_i}$ in regression loss function. 
	\begin{equation}
		L_{reg}=\sum_{i=0}^{N}w_i\times l_{reg_{i}}
		\label{con:eq4}
	\end{equation}
	The value of attention weight $w_i$ depends on the confidence score and localization precision of the related keypoint. The confidence score is the sampled value from heatmap $Y^{'}\in[0,1]^{H/4\times W/4\times C}$ via the pixel index of ground truth keypoint $p_i$. The localization precision is defined as the 3D intersection over union (3D IoU) between the detection box and its ground truth counterpart. For each keypoint, the larger the confidence score, the more attention is drawn. The closer to +1 the 3D IoU, the smaller the attention weight is. We build the attention weight equation in the form of linear combination of confidence score and 3D IoU. Then, the softmax function is utilized to obtain normalization and nonlinearity. In order to bring no significant effect on other relevant hyper-parameters in Eq.~\ref{con:eq3}, we maintain the same summation of attention weights as before by multiplying the number of keypoints. The complete formula for the calculation of attention weight $w_{i}$ is defined as Eq.~\ref{con:eq5}.
	\begin{equation}
		w_i=\frac{e^{P_i+\beta\times(1-IoU_{3D_{i}})}}{\sum_{n=0}^{N}e^{P_n+\beta \times(1-IoU_{3D_{n}})}}\times N,
		\label{con:eq5}
	\end{equation}
	where $P_i$ denotes the confidence score, $IoU_{3D_i}$ represents the 3D IoU of i-th keypoint, $\beta$ is the hyper-parameter that controls the impact ratio of $IoU_{3D_i}$ and $N$ specifies the total number of keypoints per batch images.
    
    \begin{table*}[h!t]
    \scriptsize
    \centering
    \small
    \renewcommand\arraystretch{1.0}
    \begin{center}
    \begin{tabular}{|c|c|c|c|c|c|c|c|c|c|}
    \hline
    \multirow{2}{*}{Methods}& \multirow{2}{*}{Backbones}& \multirow{2}{*}{FPS}& \multicolumn{3}{|c|}{$AP_{3D}(IOU>0.7)$}& \multicolumn{3}{|c|}{$AP_{BEV}(IOU>0.7)$} \\
    \cline{4-9}               
          &    &      &  Easy&    Moderate&   Hard&   Easy&   Moderate&  Hard\\
    \hline
    SMOKE\cite{liu2020smoke}&  ResNet-18&   86.80& 12.11& 10.64& 10.23& 16.07& 13.19& 11.68\\
    Ours-SMOKE&  ResNet-18&   88.57& 17.04& 14.02& 12.23& 23.85& 19.68& 16.91\\
    Delta&  -&   {\color{ForestGreen}+1.77}& {\color{ForestGreen}\textbf{+4.93}}& {\color{ForestGreen}+3.38}& {\color{ForestGreen}+2.00}& {\color{ForestGreen}\textbf{+7.78}}& {\color{ForestGreen}\textbf{+6.49}}& {\color{ForestGreen}\textbf{+5.23}}\\
    \hline
    \hline
    SMOKE\cite{liu2020smoke}&  ResNet-34&   69.52& 13.79& 11.85& 11.23& 18.85& 15.89& 14.64\\
    Ours-SMOKE&  ResNet-34&   71.32& 18.01& 15.29& 14.28& 24.83& 20.50& 17.65 \\ 
    Delta&  -&   {\color{ForestGreen}+1.80}& {\color{ForestGreen}+4.22}& {\color{ForestGreen}\textbf{+3.44}}& {\color{ForestGreen}+3.05}& {\color{ForestGreen}+5.98}& {\color{ForestGreen}+4.61}& {\color{ForestGreen}+3.01}\\
    \hline
    \hline
    SMOKE\cite{liu2020smoke}&  DLA-34&   40.32& 14.76& 12.85& 11.50& 19.99& 15.61& 15.28\\
    Ours-SMOKE&  DLA-34&   42.37& 19.31& 16.19& 15.47& 25.45& 21.22& 17.91\\
    Delta&       -&   {\color{ForestGreen}\textbf{+2.05}}& {\color{ForestGreen}+4.55}& {\color{ForestGreen}+3.34}& {\color{ForestGreen}\textbf{+3.97}}& {\color{ForestGreen}+5.46}& {\color{ForestGreen}+5.61}& {\color{ForestGreen}+2.63}\\
    \hline
    \end{tabular}
    \end{center}
    \caption{Performance comparison of SMOKE and its improved model with our proposed methods on the KITTI val 3D and BEV detection benchmark. Both metrics for car category are evaluated by $AP|_{R_{11}}$ at 0.7 IoU threshold. The frame rate is measured on a RTX 2080Ti.}
    \label{table:tab1}
    \end{table*}
    
    \begin{table*}[h!t]
    \scriptsize
    \centering
    \small
    \renewcommand\arraystretch{1.0}
    \begin{center}
    \begin{tabular}{|c|c|c|c|c|c|c|c|c|c|}
    \hline
    \multirow{2}{*}{Methods}& \multirow{2}{*}{Backbones}& \multirow{2}{*}{GPU}& \multirow{2}{*}{FPS}& \multicolumn{3}{|c|}{$AP_{3D}(IOU>0.7)$}& \multicolumn{3}{|c|}{$AP_{BEV}(IOU>0.7)$} \\
    \cline{5-10}               
          &    &      &      &  Easy&    Moderate&   Hard&   Easy&   Moderate&  Hard\\
    \hline
    FQNet\cite{liu2019deep}      & VGG-16&  1080Ti & 2.00& 2.77& 1.51& 1.01& 5.40& 3.23& 2.46\\
    MonoGRNet\cite{9409679}& VGG-16& Tesla P40& 16.67& 9.61& 5.74& 4.25& 18.19& 11.17& 8.73\\
    MonoPSR\cite{Ku2019Monocular3O}& ResNet-101& Titan X& 8.34& 10.76& 7.25& 5.85& 18.33& 12.58& 9.91\\
    MonoDIS\cite{simonelli2019disentangling}& ResNet-34&  Tesla V100& 10.00& 10.37& 7.94& 6.40& 17.23& 13.19& 11.12\\
    UR3D\cite{shi2020distance}   & ResNet-34& Titan X& 8.33& \textbf{15.58}& 8.61& 6.00& 21.85& 12.51& 9.20\\
    M3D-RPN\cite{brazil2019m3d}  & DenseNet-121&1080Ti& 6.25& 14.76& 9.71& 7.42& 21.02& 13.67& 10.23\\
    MonoPair\cite{chen2020monopair}& DLA-34& 1080Ti& 17.54& 13.04& 9.99& 8.65& 19.28& 14.83& 12.89\\
    RTM3D\cite{li2020rtm3d}& DLA-34& 1080Ti& 18.18& 14.41& 10.34& \textbf{8.77}& 19.17& 14.20& 11.99\\
    \hline
    SMOKE\cite{liu2020smoke}&  DLA-34&  2080Ti&   40.32& 14.03& 9.76& 7.84& 20.83& 14.49& 12.75\\
    Ours-SMOKE&  ResNet-34   &  2080Ti&  \textbf{71.32}& 15.32& \textbf{10.64}& 8.59& \textbf{26.67}& \textbf{17.58}& \textbf{14.51}\\
    Delta&     -&  -&  {\color{ForestGreen}+31.00}& {\color{ForestGreen}+1.29}& {\color{ForestGreen}+0.88}& {\color{ForestGreen}+0.75}& {\color{ForestGreen}+5.84}& {\color{ForestGreen}+3.09}& {\color{ForestGreen}+1.76}\\
    \hline
    \end{tabular}
    \end{center}
    \caption{Results on the KITTI test 3D and BEV detection benchmark. Both metrics for car category are evaluated by $AP|_{R_{40}}$ at 0.7 IoU threshold.}
    \label{table:tab3}
    \end{table*}
    
    \begin{table}[h!t]
    \scriptsize
    \centering
    \small
    \renewcommand\arraystretch{1.0}
    \begin{center}
    \begin{tabular}{|c|c|c|c|c|c|}
    \hline
    \multirow{2}{*}{Methods}&\multirow{2}{*}{Backbones}& \multirow{2}{*}{FPS}& \multicolumn{3}{|c|}{$AP_{BEV}(IOU>0.5)$} \\
    \cline{4-6}
    &   &   & Easy&   Moderate&   Hard\\
    \hline
    \hline
    CenterNet& ResNet-18& 70.12& 33.41& 26.21& 21.74\\
    Ours& ResNet-18& 79.87& 42.58& 31.54& 28.95\\
    Delta& -& {\color{ForestGreen}\textbf{+9.75}}& {\color{ForestGreen}\textbf{+9.71}}& {\color{ForestGreen}+5.33}& {\color{ForestGreen}\textbf{+7.21}}\\
    \hline
    \hline
    CenterNet& ResNet-34& 61.58& 35.68& 28.33& 23.57\\
    Ours& ResNet-34& 66.58& 40.37& 34.60& 30.01\\
    Delta& -& {\color{ForestGreen}+5.00}& {\color{ForestGreen}+4.69}  & {\color{ForestGreen}\textbf{+6.27}}& {\color{ForestGreen}+6.44}\\
    \hline
    \hline
    CenterNet& DLA-34& 38.17& 34.00& 30.50& 26.80\\
    Ours& DLA-34& 41.32& 42.69& 35.75& 30.11\\
    Delta& -& {\color{ForestGreen}+3.18}& {\color{ForestGreen}+8.69}  & {\color{ForestGreen}+5.25}& {\color{ForestGreen}+3.31}\\
    \hline
    \end{tabular}
    \end{center}
    \caption{Performance comparison of CenterNet and its improved model with our proposed methods on the KITTI val BEV detection benchmark. Both metrics for car category are evaluated by $AP|_{R_{11}}$ at 0.5 IoU threshold. The frame rate is measured on a RTX 2080Ti.}
    \label{table:tab2}
    \end{table}
    
    \section{Experiments}
    \label{sec:experiments}
	Our proposed Lite-FPN and attention loss are framework independent module which can be integrated into a majority of keypoint-based monocular 3D object detectors. In the following experiments, we integrate the proposed methods into SMOKE\cite{liu2020smoke} and CenterNet\cite{zhou2019objects} and evaluate their performance on the KITTI object detection benchmark.
    
    \subsection{Datasets}
    \label{sec:dataset}
	KITTI dataset\cite{geiger2012we} provides images from four front-view cameras and point clouds from one HDL-64E LIDAR, which are synchronized in both time and space dimensions. There are 7841 samples in training set and 7518 samples in test set. Since the annotated 3D bounding boxes are not available in test set, we split the training set into 3712 training examples and 3769 validation examples as described in\cite{zhou2018voxelnet,yan2018second}. The evaluation benchmark of KITTI object detection requires to detect objects in three classes such as car, pedestrian and cyclist, which can be further divided into easy, moderate, and hard levels according to some object instance properties on the image plane such as the height of 2D bounding boxes, occlusion and truncation. In order to compare with previous studies, we mainly focus on the performance on car category.
     
    \begin{figure*}
    \begin{center}
    \includegraphics[width=1.0\textwidth]{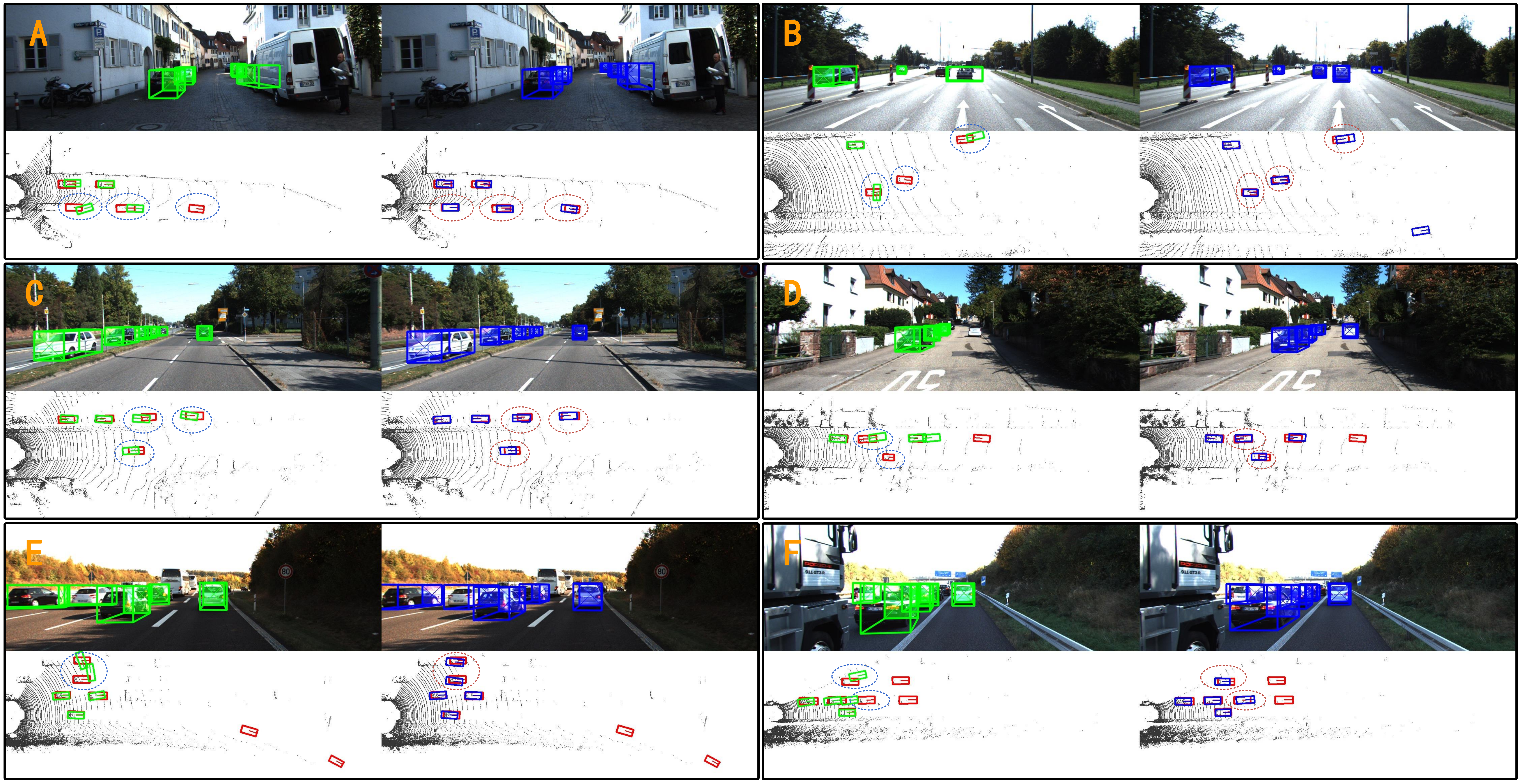}
    \end{center}
       \caption{Qualitative results on validation set of KITTI. We present six pairs of comparison figures indicated from A to F. Each pair consists of four pictures, the upper left displays the predictions of the SMOKE\cite{liu2020smoke} baseline ({\color{green}green}), the lower left is the corresponding display on bird’s-eye view, the upper right shows the results of our improved model ({\color{blue}blue}), the lower right is its bird's-eye view display. The orientation of boxes is shown by the non-transparent front part of projected 3D boxes on the image plane and the line connecting the center point and front side midpoint of 2D rotated boxes on bird's-eye view. Boxes in red color are ground truth. The dotted ellipses highlight the instances which are detected more accurately by our improved model.}
    \label{fig:four}
    \end{figure*}
     
    \begin{table*}[h!t]
    \scriptsize
    \centering
    \small
    \renewcommand\arraystretch{1.0}
    \begin{center}
    \begin{tabular}{|c|c|c|c|c|c|c|c|c|}
    \hline
    \multirow{2}{*}{Lite-FPN}& \multirow{2}{*}{Attention Loss}& \multirow{2}{*}{FPS}& \multicolumn{3}{|c|}{$AP_{3D}(IOU>0.7)$}& \multicolumn{3}{|c|}{$AP_{BEV}(IOU>0.7)$} \\
    \cline{4-9}               
                    &      &      &  Easy&    Moderate&   Hard&   Easy&   Moderate&  Hard\\
    \hline
    &               & 40.32& 14.76& 12.85& 11.50& 19.99& 15.61& 15.28\\
    $\surd$&        & 42.37& 17.37& 14.69& 14.09& 22.42& 18.87& 16.48\\
    $\surd$& $\surd$& \textbf{42.37}& \textbf{19.31}& \textbf{16.19}& \textbf{15.47}& \textbf{25.45}& \textbf{21.22}& \textbf{17.91}\\
    \hline
    \end{tabular}
    \end{center}
    \caption{Ablation study of SMOKE with different configurations on the KITTI datasets using val split. We use DLA-34 as the backbone. The frame rate is measured on a RTX 2080Ti and the best results are highlighted in bold type.}
    \label{table:tab4}
    \end{table*}
    
    \subsection{Implementation Details}
    \label{sec:implementation_details}	
	{\bf CenterNet.} CenterNet\cite{zhou2019objects}, as a seminal work of one-stage keypoint-based detectors, uses keypoint estimation to detect the center point of 2D box and directly regresses all other object properties, such as depth, dimension, orientation and the joint locations of human pose at each keypoint location. Based on the official code\footnote{https://github.com/xingyizhou/CenterNet} released by author, we merge the existing five regression branches into two new heads: a 2D box head responsible for keypoint local offset and 2D size regression, a 3D box head in charge of predicting depth, orientation and 3D size. These two heads are composed of the same layers as the original regression branch. Then, two Lite-FPN modules are applied on 2D box and 3D box heads. Our  proposed  attention loss is chosen to build the regression loss. Besides, we change the output channels of three deconvolutional layers in backbone to 256, 128 and 64 respectively. The hyper-parameter $\beta$ in Eq.~\ref{con:eq5} is 0.25. We maintain nearly the same training strategy as the baseline except for a batch size of 8.
	
	{\bf SMOKE.} The 3D detection performance of CenterNet\cite{zhou2019objects} is restricted by the distinction between 2D center point and projected 3D center point on the image space. SMOKE\cite{liu2020smoke} resolves this inherent drawback by eliminating the 2D detection branch and directly estimating the projected 3D center point. The official released code\footnote{https://github.com/lzccccc/SMOKE} is used for our comparison experiments. Based on the baseline code base, we replace the final convolutional layer in regression branch with our proposed Lite-FPN module and displace the regression loss with our proposed attention loss. The  hyper-parameter $\beta$ in Eq.~\ref{con:eq5} is set to 0.5. We follow nearly the same training strategy as the baseline except for a batch size of 8.
    
    \subsection{Quantitative Analysis}
    \label{sec:quantitative_analysis}
	{\bf Validation Set.} As shown in Table.~\ref{table:tab1}, the improved SMOKE\cite{liu2020smoke} with ResNet-18, ResNet-34 and DLA-34 backbones outperforms the baseline with the same backbones by 3.38\%, 3.44\% and 3.34\% in $AP_{3D}$ on moderate level respectively. Furthermore, the improved SMOKE\cite{liu2020smoke} with ResNet-18 backbone also exceeds the $AP_{3D}$ metric of the baseline with DLA-34 backbone by 1.17\%, while the frame rate is more than twice that of the latter. 
	
	The comparative experiments of CenterNet\cite{zhou2019objects} are shown in Table.~\ref{table:tab2}. We choose $AP_{BEV}$ at an IoU of 0.5 as the evaluation metric which is consistent with the CenterNet paper. The improved CenterNet\cite{zhou2019objects} with ResNet-18, ResNet-34 and DLA-34 backbones achieves the performance improvements of 5.33\%, 6.27\% and 5.25\% on moderate level over the baseline with the same backbones.
    
    {\bf Test Set.} We compare the improved SMOKE\cite{liu2020smoke} with other state-of-the-art detectors on the official KITTI test set. As shown in Table.~\ref{table:tab3}, our improved model with ResNet-34 backbone reaches 71.32 FPS on 2080Ti and achieves the best speed-accuracy trade-off.
	
	Based on the analysis above, our proposed methods can improve performance and meanwhile reduce inference time, which is appropriate for the edge computing platform in autonomous driving.
    
    \subsection{Qualitative Analysis} 
    \label{sec:qualitative_analysis}
	Qualitative results of SMOKE\cite{liu2020smoke} with DLA-34 backbone and its improved model integrating our proposed methods are displayed in Figure.~\ref{fig:four}. All sensor data are from the validation set of KITTI. For better visualization and comparison, we visualize detections in the form of projected 3D boxes on the image plane and 2D rotated boxes on bird's-eye view. Point clouds are only for visualization purpose. The comparative figures demonstrate that our contributions can improve localization precision (A-F in Figure.~\ref{fig:four}) and reduce miss detection (A-C in Figure.~\ref{fig:four}) effectively.
    
    \subsection{Ablation Studies}
    \label{sec:ablation_studies}
	We conduct ablation experiments to examine how our proposed Lite-FPN module and attention loss affect the performance of keypoint-based detectors. We choose the SMOKE\cite{liu2020smoke} with DLA-34 backbone as baseline and add the proposed methods one-by-one. All ablations are evaluated on KITTI 3D and BEV detection benchmark using val split. From the results listed in Table.~\ref{table:tab4}, some promising conclusions can be summed up as follows:\\
	{\bf Lite-FPN module is crucial.} The results of “+Lite-FPN” outperform the baseline by 1.84\% in $AP_{3D}$ and 3.26\% in $AP_{BEV}$ on moderate level with a higher frame rate, which indicates that our proposed Lite-FPN module manages to improve algorithm performance in an efficient way. \\
	{\bf Attention loss is promising.} The combination of Lite-FPN and attention loss improves the performance of “+Lite-FPN” by 1.50\% and 2.35\% in terms of $AP_{3D}$ and $AP_{BEV}$ on moderate level respectively, which verifies the effectiveness of our proposed attention loss.
    
    \section{Conclusion and Future Work}
    \label{sec:conclusion_and_future_work}
	In this paper, to enhance the performance of keypoint-based monocular 3D object detectors, we propose a generic Lite-FPN module which is beneficial to handle objects within a large range of scales and distances. We further present a novel attention loss to alleviate the misalignment between classification score and localization precision. Comparative experiments on the KITTI dataset demonstrate that the improved models with our proposed methods can achieve much higher accuracy and lower latency than the baselines. In the future, we aim to extend Lite-FPN module in terms of improving the feature fusion method and eliminating the discretization error caused during the process of generating the pixel indices of keypoints on lower resolution feature maps. 
    
    \bibliography{egpaper_final}
    \bibliographystyle{ieee_fullname}
\end{document}